\documentclass[10pt,letterpaper]{article}
\usepackage[T1]{fontenc}
\usepackage[utf8]{inputenc}
\usepackage{lmodern,microtype}
\usepackage[margin=0.94in,top=0.82in,bottom=0.86in]{geometry}
\usepackage{amsmath,booktabs,array,graphicx,xcolor,tikz}
\usetikzlibrary{arrows.meta}
\usepackage[colorlinks=true,linkcolor=black,citecolor=black,urlcolor=blue!40!black]{hyperref}
\hypersetup{pdftitle={MoP: Mixture of Predictors},pdfauthor={Bohang Sun, Max Zhu, Pietro Liò, Jialin Yu}}
\newcommand{\pair}[2]{\shortstack[r]{#1\\[2pt]#2}}
\newcommand{\nhead}[2]{\shortstack[c]{#1\\#2}}
\newcommand{\work}[2]{\href{#1}{\emph{#2}}}
\definecolor{ink}{HTML}{24435D}
\definecolor{fillprefix}{HTML}{EAF0F5}
\definecolor{fillcheap}{HTML}{E8F0E8}
\definecolor{fillfull}{HTML}{F6EBDD}

\title{\bfseries MoP: Mixture of Predictors}
\author{Bohang Sun$^{1}$ \thanks{\texttt{bs835@cam.ac.uk}.}\quad Max Zhu$^{2}$ \quad
  Pietro Liò$^{1}$\thanks{\texttt{pl219@cam.ac.uk}.} \quad
  Jialin Yu$^{3,4}$\thanks{\texttt{yu.jialin@outlook.com}.}\\[0.6em]
  {\small $^{1}$Department of Computer Science and Technology, University of Cambridge}\\
  {\small $^{2}$Department of Pure Mathematics and Mathematical Statistics, University of Cambridge}\\
  {\small $^{3}$Department of Engineering Science, University of Oxford}\\
  {\small $^{4}$University of Bath}}
\date{}

\begin{document}
\maketitle
\vspace{-1.5em}
\begin{abstract}
Cross-layer key--value sharing lets a language model construct the history of a sequence before its final layers predict the next token. We use this separation to reduce the average cost of generating tokens. In \emph{Mixture of Predictors} (MoP), the original prefix updates the shared state at every step. A learned router selects positions for a distilled two-layer decoder with a 512-token attention window, bypassing the full decoder. Low confidence predictions by the small decoder are rerouted back to the full decoder, which makes predictions together with the non-selected tokens. We evaluate four pretrained models on 1,260 questions from five benchmarks. At 512--800 resident requests on GH200 GPUs, the short decoder supplies 36.5--48.9\% of decoded tokens and throughput rises by 13.7--18.2\% over each model's full-depth decoder. This speedup comes with minimal accuracy penalty, differences range from $-0.64$ to $+2.13$ percentage points. On Bolt, for example, throughput rises from 11,622 to 13,222 tokens/s at 600 requests, and the two systems answer 676 and 675 of their 1,253 jointly completed questions correctly.
\end{abstract}

\begin{figure}[htbp]
\centering
\includegraphics[width=\linewidth]{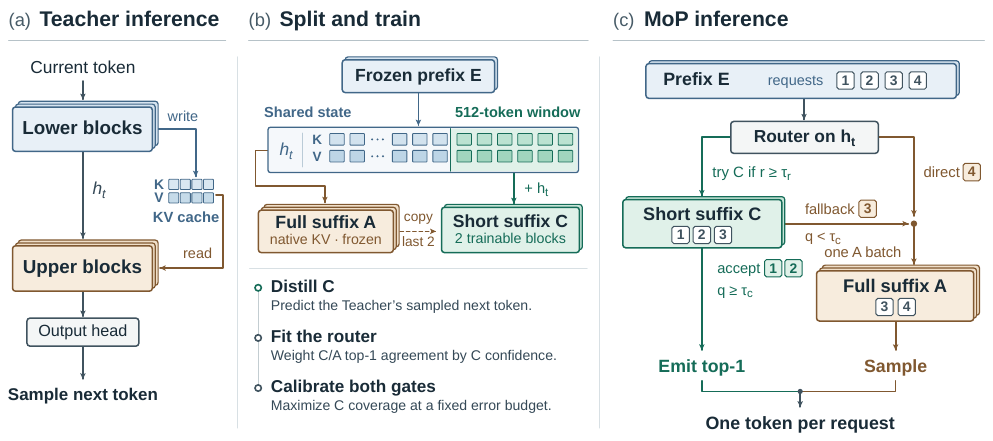}
\caption{\textbf{MoP inference and training.} (a) The Teacher updates its shared history with prefix $E$ and samples with full suffix $A$. (b) The last two blocks of $A$ initialize $C$, which learns from Teacher-generated tokens using a 512-token attention window. (c) Requests 1--3 pass the router gate $r\ge\tau_r$. The short suffix emits its top token for requests 1 and 2, where $q=\max_v p_C(v)\ge\tau_c$. Request 3 falls back and joins request 4 in a single call to $A$.}
\label{fig:overview}
\end{figure}

\section{Introduction}
An autoregressive model does two things when it reads a token. It writes information that later positions can attend to, and it computes a distribution for the next token. A conventional Transformer interleaves these operations throughout its depth. In YOCO, one stack builds a shared key--value (KV) cache and a second reads it \cite{yoco}. SwiftKV exploited a similar separation for enterprise workloads with roughly ten input tokens per output token. Computing later-layer KV from an earlier representation allowed most prompt positions to skip the upper layers \cite{swiftkv}. Recent architectures use this idea in different ways. KITE dedicates one tower to producing KV, while DepthGraft reconstructs deep KV from multiple earlier layers \cite{kite,depthgraft}.

Generation also benefits from this division of labor. A reasoning trace can contain thousands of output tokens, each requiring a prediction. In our Bolt evaluation, 2.65 million input tokens lead to 3.79 million first-pass Teacher output tokens. If the prefix updates the shared history on every step, a shorter suffix can predict some of those tokens. When the full suffix resumes, it has access to everything the shorter suffix generated.

We introduce \emph{Mixture of Predictors} (MoP) for pretrained models with this property. We retain the original prefix and suffix, distill a two-layer alternative on Teacher generations, and learn when to consult it (Figure~\ref{fig:overview}). An accepted prediction replaces a suffix of 14, 18 or 40 blocks with two. Under continuous batching, however, different requests select different paths, and a rejected proposal incurs the cost of both. We bring all requests that require the original suffix into one batch per round. We measure the resulting speed and answer accuracy against the full-depth Teacher at matched concurrency and precision.

\section{Method}
\subsection{A shared history and two prediction paths}
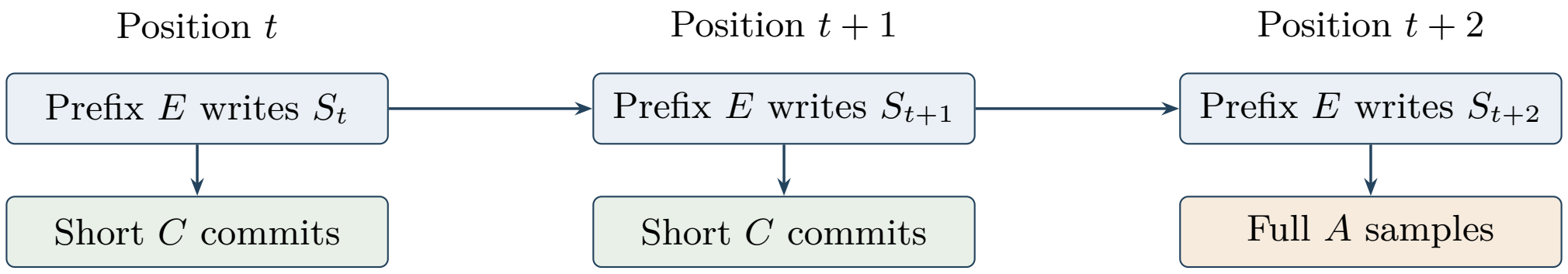
\begin{figure}[ht]
\centering
\begin{tikzpicture}[x=1cm,y=1cm,font=\small,
  box/.style={draw=ink,rounded corners=2pt,minimum width=3.25cm,minimum height=.60cm,inner sep=3pt},
  flow/.style={-{Stealth[length=4.3pt]},line width=.65pt,draw=ink}]
  \foreach \x/\k/\s in {0/{t}/{S_t},5/{t+1}/{S_{t+1}},10/{t+2}/{S_{t+2}}}{
    \node[font=\small] at (\x,1.48) {Position $\k$};
    \node[box,fill=fillprefix] (p\x) at (\x,.77) {Prefix $E$ writes $\s$};
  }
  \node[box,fill=fillcheap] (c0) at (0,-.28) {Short $C$ commits};
  \node[box,fill=fillcheap] (c5) at (5,-.28) {Short $C$ commits};
  \node[box,fill=fillfull] (a10) at (10,-.28) {Full $A$ samples};
  \draw[flow] (p0)--(p5); \draw[flow] (p5)--(p10);
  \draw[flow] (p0)--(c0); \draw[flow] (p5)--(c5); \draw[flow] (p10)--(a10);
\end{tikzpicture}
\caption{Three successive decode decisions for one request. A committed token becomes the next input to $E$. At the third position $A$ reads the history that includes both tokens emitted by $C$.}
\label{fig:state}
\end{figure}

Let $E$ be the original prefix and $A$ the original prediction suffix. On reading token $x_t$, $E$ updates the persistent state $S_t$ and produces a representation $h_t$. The model family determines the contents of $S_t$. It includes attention KV and, in Phi, the recurrent state and the memory readout used by the upper layers. The split is valid when the state needed at position $t+1$ can be advanced by $E$ from the tokens already committed. Both suffixes read this state. Neither path writes branch-specific history required for a later call to $A$ (Figure~\ref{fig:state}).

The Teacher always runs $E$ and $A$. In MoP, a small MLP reads $h_t$ and outputs a score $r_t$. A score below threshold $\tau_r$ sends the row directly to $A$. For the remaining rows, $C$ applies the shared output head and a unit-temperature softmax over the full vocabulary to obtain $p_C$. Its attention reads at most 512 KV positions, including the current position. We write $S_t^{(512)}$ for this view, including the current recurrent readout when present. The prefix representations can carry information from earlier tokens. If $\max_v p_C(v)\geq\tau_c$, MoP commits $\arg\max_v p_C(v)$ immediately. Otherwise $A$ samples the token. The first output token always comes from the native Teacher prefill.

\subsection{Teacher trajectories and suffix distillation}
Each backbone generates its own training corpus. From a general instruction pool \cite{tulu} we assign 65,000 prompts to suffix training and 10,000 separate prompts to router training. The Teacher generates an answer to each prompt. We save the prompt and answer token IDs so the suffix and router can later see exactly the same prefix at any answer position. We filter dominant or sustained exact repetitions and use a separate validation split to select checkpoints. The cleaned general suffix pools contain 64,931 trajectories for Gemma, 64,970 for Phi, 63,352 for SwiftKV and 64,989 for Bolt. For code specialization we collect additional Teacher completions on TACO training prompts \cite{taco}, checking the final code block, Python syntax and the required entry point.

We copy the Teacher's last two blocks to initialize $C$ and freeze the backbone and shared output head. For each trajectory we replay its tokens in causal order and train on supervised answer positions $\mathcal I$ using
\[
  \mathcal L_C=-\frac{1}{|\mathcal I|}\sum_{t\in\mathcal I}
  \log p_C(x_{t+1}\mid h_t,S_t^{(512)}) .
\]
The target $x_{t+1}$ is the recorded next token, including when the Teacher sampled it. The 512-position attention window is active during training. Gemma and Bolt preserve the copied local block and restrict the global block. Phi's gated memory block reads the current prefix readout and its attention block uses the short window. Both SwiftKV blocks use the window. Gemma, Phi and Bolt also receive code-specialization updates mixed with general trajectories. In Phi's final training stage, only positions with Teacher confidence above 0.8 contribute to the loss.

\subsection{Router training and choosing the thresholds}
We replay the separate router trajectories through the frozen prefix and suffixes. At each position we record the short-suffix confidence $q_t=\max_v p_C(v)$ and whether $C$ and $A$ have the same top token on that prefix. We train a 128-hidden-unit MLP on prefix features using binary cross-entropy with soft target $q_t\,\mathbf{1}[\arg\max p_C=\arg\max p_A]$. Gemma reuses its router after code specialization and recalibrates the gates. Phi fine-tunes its router against the selected short suffix. The deployed $C$/router checkpoints are 900/600, 458/500, 1,900/608 and 920/2,504 training steps for Gemma, Phi, SwiftKV and Bolt, respectively.

To select $(\tau_r,\tau_c)$, we replay two archived Teacher answers per prompt on an 800-prompt calibration set. We compute the router score, $C$ confidence and top-token agreement once per position, then evaluate candidate threshold pairs on those scores. With $U$ the accepted positions and $N$ the total number scored, we maximize $|U|/N$ subject to
\[
  \frac{1}{N}\sum_{t\in U}\mathbf{1}\!\left[\arg\max p_C(\cdot\mid x_{\leq t})
             \ne \arg\max p_A(\cdot\mid x_{\leq t})\right]\leq\epsilon .
\]
Ties favor fewer rejected attempts. Gemma and Bolt use $\epsilon=0.1\%$ throughout, with respective threshold pairs (rounded to five decimal places) $(0.53959,0.99028)$ and $(0.27474,0.98705)$. SwiftKV uses $0.2\%$ and $(0.60276,0.99331)$. Phi uses $0.3\%$ during reasoning, $(0.47615,0.98103)$, and $0.2\%$ in the final answer, $(0.48917,0.98724)$. Each reported rate counts individual token positions. These thresholds control top-token disagreement on the calibration prefixes. The accepted branch's deterministic choice changes the Teacher sampling distribution.

On all calibration positions, $C$ and $A$ agree on the top token 84.30\% of the time for Gemma, 80.63\% for Phi, 79.94\% for SwiftKV and 84.21\% for Bolt. On accepted positions, agreement is 99.78\% for Gemma, 99.47\% for SwiftKV and 99.79\% for Bolt. Applying Phi's reasoning and final-answer thresholds to the same archived positions gives 99.30\% and 99.50\%, respectively.

\subsection{Continuous batching}
Each decode round runs $E$ once for the active requests, dispatches the rows selected for $C$, and combines the rejected rows with those routed directly to $A$. The combined batch calls $A$ once. Thus every active request commits one token in that round. We use paged KV storage, CUDA Graphs bucketed by context length and batch width, and choose $A$ graph widths by timing representative groups of live rows. Padding copies the shortest valid history in the group. On code problems, the scheduler routes tokens inside a fenced code block in the final answer to $A$, including fences without a language tag. The transition to the final answer follows the model's generated delimiter in Gemma, Bolt and Phi. SwiftKV starts in the answer phase.

\section{Evaluation}
\label{sec:results}
\paragraph{Models and workload.}
Table~\ref{tab:main} compares MoP with its Teacher on Gemma 4 E4B-IT \cite{gemma}, Phi-4-mini-flash-reasoning \cite{phi}, Llama-3.3-SwiftKV-70B-Instruct \cite{swiftkv}, and NVIDIA Nemotron 3.5 Bolt E4B \cite{bolt}. Bolt is post-trained from Gemma 4 E4B-IT. The prefix/suffix splits, in blocks, are 24/18, 18/14, 40/40 and 24/18, respectively. The suite contains 400 English Global-MMLU-Lite questions \cite{global}, 172 LiveCodeBench v6 problems \cite{lcb}, 30 AIME 2026 problems \cite{aime}, 460 BBEH Mini questions \cite{bbeh} and 198 GPQA Diamond questions \cite{gpqa}. Each benchmark uses a fixed set of few-shot examples from outside the evaluated questions, totaling 1,486--1,502 tokens in the Gemma tokenizer. The Global-MMLU-Lite and BBEH subsets add 860 questions to the evaluation. The remaining 400 prompt IDs also appear in the historical threshold-calibration workload. The 1,260 official IDs include two BBEH items with identical question text and different answer keys.

\begin{table}[ht]
\centering
\caption{Five-benchmark evaluation on 1,260 questions. Paired cells show \textbf{Teacher above MoP}. Task scores give correct answers over jointly completed questions. Overall accuracy weights the $N$ jointly completed questions equally. Decode throughput uses full-residency rounds. Peak GPU memory includes setup and uses the larger per-card peak for SwiftKV.}
\label{tab:main}
\vspace{4pt}
\begingroup
\fontsize{9}{11}\selectfont
\setlength{\tabcolsep}{2.2pt}
\renewcommand{\arraystretch}{1.12}
\begin{tabular*}{\linewidth}{@{\extracolsep{\fill}}lrrrrrrrcrr@{}}
\toprule
& \multicolumn{5}{c}{Correct / jointly completed}
& \multicolumn{2}{c}{Steady decode} & \multicolumn{1}{c}{Peak GPU} & \multicolumn{2}{c}{Overall}\\
\cmidrule(lr){2-6}\cmidrule(lr){7-8}\cmidrule(l){10-11}
Model & \nhead{Global}{MMLU} & \nhead{LiveCode}{Bench} & AIME & BBEH & GPQA
 & tokens/s & gain & GiB & $N$ & ACC (\%)\\
\midrule
\nhead{Gemma}{$B=600$}
& \pair{335/400}{330/400} & \pair{121/172}{116/172} & \pair{14/30}{14/30} & \pair{109/457}{119/457} & \pair{118/198}{110/198}
& \pair{11,162}{13,100} & $+17.36\%$ & \pair{77.18}{77.78} & 1,257 & \pair{55.45}{54.81}\\[4pt]
\nhead{Phi}{$B=800$}
& \pair{119/282}{122/282} & \pair{22/111}{19/111} & \pair{7/16}{8/16} & \pair{11/178}{16/178} & \pair{58/164}{68/164}
& \pair{4,378}{5,173} & $+18.15\%$ & \pair{95.54}{95.20} & 751 & \pair{28.89}{31.03}\\[4pt]
\nhead{SwiftKV}{$B=512$}
& \pair{346/398}{344/398} & \pair{60/172}{56/172} & \pair{2/30}{2/30} & \pair{85/448}{88/448} & \pair{87/195}{96/195}
& \pair{1,471}{1,673} & $+13.70\%$ & \pair{77.29}{80.71} & 1,243 & \pair{46.66}{47.14}\\[4pt]
\nhead{Bolt}{$B=600$}
& \pair{323/400}{322/400} & \pair{120/172}{119/172} & \pair{16/30}{13/30} & \pair{118/453}{123/453} & \pair{99/198}{98/198}
& \pair{11,622}{13,222} & $+13.76\%$ & \pair{77.47}{77.88} & 1,253 & \pair{53.95}{53.87}\\
\bottomrule
\end{tabular*}
\endgroup
\end{table}

\paragraph{Serving and scoring.}
Each model and its Teacher run the same ordered requests, once per arm, with seed 42 and a sampling stream indexed by request and output position. Gemma and Bolt use temperature 1.0, top-$k$ 64 and top-$p$ 0.95. Phi uses $(0.6,50,0.95)$ and SwiftKV $(0.6,50,0.90)$. These settings apply to Teacher and $A$. For Gemma, Phi and Bolt, the original suffix uses W8A16 FFN projections in both arms. SwiftKV uses W8A16 for all backbone linear layers and the shared vocabulary head, FP8 paged KV, and shared few-shot prefix pages. Batches use one GH200 per arm for Gemma, Phi and Bolt. SwiftKV partitions 40 layers on each of two GH200s.

The first pass caps total length at 16,384 for Gemma and Bolt and 8,192 for SwiftKV. Phi permits up to 16,384 new tokens within 20,736 total. Graph buckets end at 4,096, 8,192 and, where applicable, 16,384 and 20,736. After 32 warm-up rounds, we measure steady-state speed only on rounds with exactly the target number of active requests. Wall throughput includes prefill, refill and drain. We sample GPU board memory at 200\,ms intervals. For accuracy we continue each capped answer from its saved tokens and sampling position to EOS or the native context limit (131,072 for Gemma and SwiftKV, 262,144 for Phi, 65,536 for Bolt). In Teacher/MoP order, the first-pass caps affect 45/51 answers on Gemma, 207/201 on Phi, 21/16 on SwiftKV and 35/37 on Bolt. Continuation is timed separately. Phi also stops when an 8--128-token pattern repeats three times in succession. We score question IDs reaching EOS in both arms. LiveCodeBench uses public and hidden tests, BBEH its official answer matcher, and the remaining tasks their reference answers.

\paragraph{Task accuracy.}
Table~\ref{tab:main} gives the per-task results. In aggregate, Teacher and MoP answer 697 and 689 of 1,257 questions on Gemma, 217 and 233 of 751 on Phi, 580 and 586 of 1,243 on SwiftKV, and 676 and 675 of 1,253 on Bolt. The respective differences in percentage points and their paired 95\% intervals are $-0.64$ $[-2.86,1.59]$, $+2.13$ $[-0.93,5.19]$, $+0.48$ $[-1.13,2.09]$ and $-0.08$ $[-2.23,2.08]$. We resample question IDs within each benchmark 10,000 times, pairing the two arms. These intervals describe question-level variation at the evaluated seed. Phi has the lowest completion coverage. Its repetition rule fires on 325 Teacher and 333 MoP answers. Counting all 1,260 questions and assigning zero to unfinished answers gives 246 and 270 correct on Phi.

\paragraph{Throughput and executed depth.}
Full-residency throughput improves on all four models. For SwiftKV the finite page pool sustains 512 active requests for 87 measured Teacher rounds and 84 MoP rounds. With prefill, replenishment and drain included, first-pass wall throughput rises by 11.77\% on Gemma, 34.79\% on Phi, 9.98\% on SwiftKV and 5.25\% on Bolt. Output lengths can differ between arms. Bolt, for instance, generates 3.79 and 3.90 million tokens and takes 641 and 626 seconds, respectively, for the same 1,260 first-pass requests.

Let $u$ be the fraction of decode positions where $C$ emits the token and $v$ the fraction where it is attempted. If the full suffix has $a$ blocks and $C$ has two, the mean net saving in block evaluations per emitted token is $a u - 2v$. This charges both branches on a rejected attempt. Over the first pass, $(u,v)$ in percent is $(48.49,80.92)$ for Gemma, $(45.09,86.16)$ for Phi, $(36.50,75.21)$ for SwiftKV, and $(48.87,91.83)$ for Bolt. These rates yield \textbf{7.11, 4.59, 13.10 and 6.96} fewer logical block evaluations per token, respectively. GPU time also reflects attention span, graph padding, output projection and dispatch. On Bolt a full 600-row decode round falls from 51.62 to 45.38\,ms.

\section{Related work}
Cross-layer sharing comes in several forms. HySparse2 builds cross-decoder attention caches from self-decoder features \cite{hysparse}. The KITE authors discuss KV-Mirror in WeLM and implement separate KV-producing and KV-reading towers in their Step Scale Transformer \cite{kite}. DeepSeek-V4.1-Flash constructs global decoder KV in the encoder while retaining layer-local sliding-window state \cite{deepseek}. This distinction matters when choosing the MoP split. U-shaped Mixture-of-Loops also separates memory construction from prediction, using a recurrent architecture trained with shared loops \cite{molt}. MoP learns an alternative readout on top of the state interface of an existing checkpoint.

Confidence-based early exits allocate depth by token \cite{calm}. River-LLM maintains KV during early exit through a shared exit path \cite{river}. In speculative decoding the target verifies draft tokens, including in methods designed for long contexts and large batches \cite{magicdec}. In MoP a confident short-suffix prediction is final and the full suffix serves the remaining rows. This lets us study the tradeoff between task accuracy and measured throughput under continuous batching.

\section{Conclusion}
A pretrained backbone that builds its historical state before its prediction suffix can give different tokens different readout depths. Across four checkpoints the two-layer MoP suffix commits more than a third of decode tokens. Grouping all other rows into one full-suffix call per round yields 13.7--18.2\% higher steady-state decode throughput at the evaluated concurrency levels.


\begin{thebibliography}{99}\small
\setlength{\itemsep}{2pt}
\bibitem{yoco} Yutao Sun et al. \work{https://arxiv.org/abs/2405.05254}{You Only Cache Once: Decoder-Decoder Architectures for Language Models}. NeurIPS, 2024.
\bibitem{swiftkv} Aurick Qiao, Zhewei Yao, Samyam Rajbhandari, and Yuxiong He. \work{https://aclanthology.org/2025.emnlp-main.1306/}{SwiftKV: Fast Prefill-Optimized Inference with Knowledge-Preserving Model Transformation}. EMNLP, 2025. \href{https://huggingface.co/Snowflake/Llama-3.3-SwiftKV-70B-Instruct}{Llama-3.3-SwiftKV-70B-Instruct model card}.
\bibitem{tulu} Allen Institute for AI. \work{https://huggingface.co/datasets/allenai/tulu-3-sft-mixture}{T\"ulu 3 supervised fine-tuning mixture}. Dataset, 2024.
\bibitem{taco} Rongao Li et al. \work{https://arxiv.org/abs/2312.14852}{TACO: Topics in Algorithmic COde Generation Dataset}. 2023.
\bibitem{gemma} Gemma Team. \work{https://arxiv.org/abs/2607.02770}{Gemma 4 Technical Report}. 2026.
\bibitem{phi} Liliang Ren et al. \work{https://arxiv.org/abs/2507.06607}{Decoder-Hybrid-Decoder Architecture for Efficient Reasoning with Long Generation}. 2025.
\bibitem{bolt} NVIDIA. \work{https://huggingface.co/nvidia/NVIDIA-Nemotron-3.5-Bolt-E4B-BF16}{NVIDIA Nemotron 3.5 Bolt E4B BF16}. Model card, 2026.
\bibitem{global} Shivalika Singh et al. \work{https://aclanthology.org/2025.acl-long.919/}{Global MMLU: Understanding and Addressing Cultural and Linguistic Biases in Multilingual Evaluation}. ACL, 2025.
\bibitem{lcb} Naman Jain et al. \work{https://arxiv.org/abs/2403.07974}{LiveCodeBench: Holistic and Contamination Free Evaluation of Large Language Models for Code}. ICLR, 2025.
\bibitem{aime} MathArena. \work{https://huggingface.co/datasets/MathArena/aime_2026}{AIME 2026}. Dataset, 2026.
\bibitem{bbeh} Mehran Kazemi et al. \work{https://aclanthology.org/2025.acl-long.1285/}{BIG-Bench Extra Hard}. ACL, 2025.
\bibitem{gpqa} David Rein et al. \work{https://arxiv.org/abs/2311.12022}{GPQA: A Graduate-Level Google-Proof Q\&A Benchmark}. 2023.
\bibitem{hysparse} Jianyu Wei et al. \work{https://arxiv.org/abs/2609.26368}{HySparse2: Hybrid Sparse Attention with Two-Level KV Sharing}. 2026.
\bibitem{kite} Zhiheng Hu et al. \work{https://arxiv.org/abs/2609.27294}{KITE: KV-Invariant Transformer Expansion for Efficient Agentic LLM Scaling}. 2026.
\bibitem{depthgraft} Xiaohan Qin et al. \work{https://arxiv.org/abs/2611.03142}{DepthGraft: Structural Regularization through Hierarchical Cross-Layer KV Reconstruction}. 2026.
\bibitem{deepseek} DeepSeek-AI. \work{https://arxiv.org/abs/2609.19969}{DeepSeek-V4.1-Flash: Pushing the Limits of KV Cache Compression}. 2026.
\bibitem{molt} Lianghui Zhu et al. \work{https://arxiv.org/abs/2611.34923}{Mixture-of-Loops Transformer}. 2026.
\bibitem{calm} Tal Schuster et al. \work{https://arxiv.org/abs/2207.07061}{Confident Adaptive Language Modeling}. NeurIPS, 2022.
\bibitem{river} Yingtao Shen and An Zou. \work{https://aclanthology.org/2026.acl-long.1746/}{River-LLM: Large Language Model Seamless Exit Based on KV Share}. ACL, 2026.
\bibitem{magicdec} Ranajoy Sadhukhan et al. \work{https://arxiv.org/abs/2408.11049}{MagicDec: Breaking the Latency-Throughput Tradeoff for Long Context Generation with Speculative Decoding}. ICLR, 2025.
\end{thebibliography}
\end{document}